\documentclass[letterpaper, 10 pt, conference]{ieeeconf}

\IEEEoverridecommandlockouts

\usepackage{graphics}
\usepackage{epsfig}
\usepackage[slantedGreek]{mathptmx}
\DeclareFontShape{OT1}{ptm}{m}{scit}{<->ssub * ptm/m/it}{}
\usepackage{times}
\usepackage{amsmath}
\usepackage{amssymb}

\DeclareMathAlphabet{\mathcal}{OMS}{cmsy}{m}{n}

\DeclareMathOperator*{\argmin}{arg\,min}

\usepackage{booktabs}
\usepackage{array}
\usepackage{graphicx}
\usepackage[hidelinks]{hyperref}

\usepackage{xcolor}
\definecolor{figdepth}{HTML}{1F6FB5}\definecolor{figopt}{HTML}{B25E0E}

\title{
Pow3R-SLAM: Real-Time RGB-D SLAM with 3D Reconstruction Priors
}

\author{Christopher Kolios$^{1}$, Ishaan Mehta$^{1}$, Sasa Janjic$^{2}$,
        Yeganeh Bahoo$^{1}$ and Sajad Saeedi$^{3}$%
\thanks{This work has been submitted to the IEEE for possible publication. Copyright may be
        transferred without notice, after which this version may no longer be accessible.}%
\thanks{\raggedright $^{1}$Toronto Metropolitan University, Toronto, Canada.
        {\tt\small \{ckolios, ishaan.mehta, bahoo\}@torontomu.ca}}%
\thanks{\raggedright $^{2}$University of Windsor, Windsor, Canada.
        {\tt\small sasa.janjic@uwindsor.ca}}%
\thanks{\raggedright $^{3}$University College London, London, United Kingdom.
        {\tt\small s.saeedi@ucl.ac.uk}}%
}
\begin{document}

\maketitle
\thispagestyle{empty}
\pagestyle{empty}

\begin{abstract}

We present Pow3R-SLAM, a real-time RGB-D simultaneous localization and mapping (SLAM) system that uses Pow3R for tracking and mapping. Inspired by MASt3R-SLAM, a recent work on monocular SLAM using two-view 3D reconstruction priors,
we extend the work to incorporate depth as a prior on the network's prediction, rather than as geometry to fuse. Where traditional RGB-D SLAM systems struggle with sparsity in the depth images, Pow3R utilizes
the available depth to give a better-conditioned pointmap, while inferring the depths in empty regions from the two-view photometric, depth, and intrinsic data. Evaluated against MASt3R-SLAM following its protocol on 24 sequences from TUM, 7-Scenes, and Replica, Pow3R-SLAM runs 1.6$\times$ faster in wall time, has 15$\%$ lower mean trajectory error, a 3.1$\times$ lower unscaled error, and produces denser maps, with a 30$\%$ lower Chamfer distance. We also introduce a hybrid variant that runs 2.1$\times$ faster than MASt3R-SLAM at 25.3 frames per second (FPS), while maintaining improved tracking and mapping accuracy. Against ORB-SLAM3 in RGB-D mode, Pow3R-SLAM is more accurate on TUM, 7-Scenes, and ETH3D-SLAM, and completes every TUM sequence. While Pow3R-SLAM can struggle on a small set of self-similar scenes, its overall performance shows that adding depth as a prior for two-view 3D reconstruction SLAM can be beneficial. A project webpage is available at: \mbox{\url{https://ChrisKolios.github.io/Pow3R-SLAM}}, and code will be made open-source upon acceptance.

\end{abstract}

\section{Introduction}
Simultaneous localization and mapping (SLAM), the problem of mapping a scene while jointly localizing the input sensor relative to the map, is crucial for autonomous robots, augmented reality systems, and self-driving cars, all of which must map and track unknown environments. However, visual SLAM, which relies solely on cameras as input, can fail on visually similar but geometrically distinct scenes, and cannot observe metric scale without a reference. RGB-D SLAM adds a depth sensor so the map and pose can be optimized against both photometric and geometric information.

\begin{figure}[t!]
    \centering
    \includegraphics[width=0.98\linewidth]{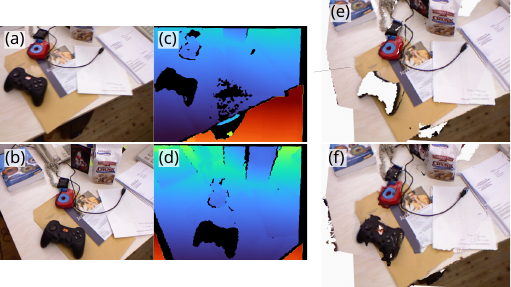}
    \caption{\textbf{Sensor depth vs. Pow3R output}. Given input frames ((a), (b)), with corresponding depth maps ((c), (d)) from the TUM fr1/desk scene~\cite{sturm2012benchmark-29f}, (e) shows the pointmap result from the back-projected depthmaps, and (f) shows the pointmap result from passing the input RGB-D into Pow3R's network. Pow3R maintains flat surfaces, and reconstructs the controller where the Kinect sensor lacks the data to do so (fills holes and regularizes surfaces).}
    \label{fig:Depthfig}
    \vspace{-0.2in}
\end{figure}

Classical depth-based SLAM systems such as KinectFusion~\cite{newcombe2011kinectfusion:-793} and RGB-D systems such as ElasticFusion~\cite{Whelan-RSS-15} treat the depth as geometry, fusing it into the map and tracking against the fused surface. However, real-world depth images are not a perfect representation of the true scene geometry. Depth sensors leave holes on absorptive, dark or shiny surfaces, and are inconsistent at object boundaries (Fig.~\ref{fig:Depthfig} (c, d)). Systems that fuse these depths as geometry inherit all of these defects, but systems that do not incorporate depth discard a key metric measurement.

Two-view 3D reconstruction prior works, including DUSt3R~\cite{wang2024dust3r:-980}, MASt3R~\cite{leroy2024grounding-709}, and Pow3R~\cite{jang2025pow3r:-27b} have shown that vision transformers (ViTs) can learn the relative geometry of two images. When passing in a pair of images, these networks can output a set of pointmaps in a common frame of reference, from which the relative poses of the cameras that took those images can be recovered. In particular, Pow3R demonstrated that passing depth, camera intrinsics, and relative pose (or any subset thereof) can condition the network, offering more accurate and geometrically consistent surfaces than RGB input alone, and inferring depth where none is provided. In Fig.~\ref{fig:Depthfig}, the back-projected depth (e) cannot fill in the controller, but Pow3R's pointmap (f) can.

Pow3R-SLAM builds on MASt3R-SLAM~\cite{murai2025mast3r-slam:-690}, a monocular system that tracks and maps with MASt3R's pointmaps, and extends it to a full RGB-D system around Pow3R. Sensor depth enters at three separable sites: conditioning the network's prediction, fixing the metric scale of every pointmap, and anchoring keyframe geometry inside the global optimization. The map is always the network's output, so holes are filled and surfaces regularized as in Fig.~\ref{fig:Depthfig}(f).

\begin{figure*}[ht]
    \centering
    \includegraphics[width=0.98\linewidth]{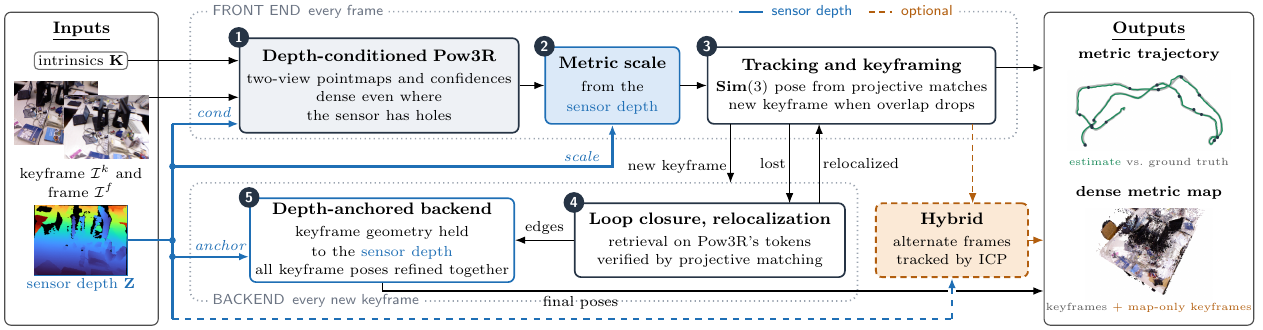}
    \caption{\textbf{Pow3R-SLAM pipeline.} Pow3R-SLAM is an RGB-D SLAM system that uses sensor depth as a prior on a two-view pointmap network. The front end (1-3) runs on every frame. (1) Pow3R predicts pointmaps and confidences for the frame and its keyframe, conditioned on sensor depth and intrinsics (Sec.~\ref{sec:depth_cal}). (2) Each prediction is made metric against the sensor depth (Eq.~\ref{eq:scaling}). (3) The frame is tracked in $\mathbf{Sim}(3)$ and may become a keyframe (Secs.~\ref{sec:matching}--\ref{sec:tracking}). The backend (4, 5) runs on every new keyframe. (4) Loop closure on Pow3R's encoder tokens adds edges in familiar locations, and relocalization locates frames the tracker has lost. (5) A global optimization refines every keyframe pose, with keyframe depths anchored to the sensor and confidence-calibrated edges (Sec.~\ref{sec:backend}, Eqs.~\ref{eq:err_back_proj}--\ref{eq:confs}). \textcolor{figdepth}{\textbf{Blue}}: the sensor-depth path. \textcolor{figopt}{\textbf{Orange}}: optional components, including the hybrid variant, which tracks alternate frames by ICP instead of Pow3R (Sec.~\ref{sec:icp}), and map-only keyframes (Sec.~\ref{sec:maponlykf}). ICP frames never become keyframes, so every keyframe comes from a Pow3R pass. Inputs and outputs: \textit{TUM fr1/desk}.}
    \label{fig:pipeline}
\end{figure*}

Pow3R-SLAM adopts MASt3R-SLAM's projective matching and second-order backend formulation, adapted to new inputs. Since Pow3R has no matching head, loop retrieval and per-correspondence confidence are rebuilt from its encoder tokens and confidence head, and map-only keyframes densify the map. A hybrid variant tracks alternate frames by iterative closest point (ICP) on the sensor depth, halving the network cost.

Our main contributions are:
\begin{itemize}
    \item To the best of our knowledge, the first RGB-D SLAM system in which sensor depth is a prior on a two-view pointmap network, entering at three separable sites.
    \item Loop retrieval and per-correspondence confidence rebuilt from Pow3R's encoder and confidence head, plus map-only keyframes that densify the map, and a hybrid ICP front end mode.
    \item Evaluation on four benchmarks against MASt3R-SLAM and ORB-SLAM3 on accuracy, reconstruction, runtime, per-site ablations, and sensitivity to depth quality.
\end{itemize}

\section{Related Work}

\textbf{Monocular SLAM} uses RGB data from a single source as input to the system. Feature-based systems such as ORB-SLAM~\cite{mur-artal2015orb-slam:-3f2} estimate poses from sparse feature correspondences across views. Dense monocular SLAM algorithms such as DTAM~\cite{newcombe2011dtam:-406} use the entire image to create a dense model of the scene. These methods are determined only up to a similarity transform, leaving metric scale unknown.

\textbf{RGB-D SLAM} systems add a depth sensor to the camera, or use depth alone. KinectFusion~\cite{newcombe2011kinectfusion:-793} fuses depth into a truncated signed distance field and tracks against it. ElasticFusion~\cite{Whelan-RSS-15} uses both RGB and depth images to jointly minimize photometric and geometric pose errors for tracking. ORB-SLAM2~\cite{mur-artal2016orb-slam2:-44b} extends ORB-SLAM's pipeline to depth, and ORB-SLAM3~\cite{campos2021orb-slam3:-eb8} adds inertial measurement unit (IMU) data and multi-map support. More modern systems such as DROID-SLAM~\cite{35402613541527} pair optical flow estimation with bundle adjustment to jointly optimize pose and dense per-pixel depth. Neural map representations have recently been popularized, with the map stored as a network or optimizable set of primitives. iMAP~\cite{sucar2021imap:-1ff} and NICE-SLAM~\cite{zhu2022nice-slam:-578} optimize implicit fields against depth, and GS-SLAM~\cite{yan2024gs-slam:-ec4} optimizes a scene of 3D Gaussians, at a cost in frame rate. All of these works consume depth as geometry to fuse, or use it as a supervision target.

A recent class of ViTs called two-view \textbf{3D reconstruction priors} takes two images and outputs dense aligned 3D point clouds. These common-frame point clouds can be used to recover camera pose (such as in DUSt3R~\cite{wang2024dust3r:-980}). MASt3R~\cite{leroy2024grounding-709}, a successor work, added per-pixel features for alignment, which improved pose recovery. Pow3R~\cite{jang2025pow3r:-27b}, a further successor, accepts any subset of depth, camera intrinsics, and relative pose, as optional inputs, which further improves the resulting point cloud. Extending this beyond pairs, \textbf{feed-forward scene models} take all views at once, such as VGGT~\cite{wang2025vggt:-709}, which predicts poses and depth for a whole set of frames.

Recent \textbf{SLAM with reconstruction priors} uses either two-view or scene prior models to track and map. MASt3R-SLAM~\cite{murai2025mast3r-slam:-690} uses MASt3R for real-time tracking and mapping by optimizing projective matches, with a \textbf{Sim}(3) pose graph and an aggregated selective match kernel (ASMK)~\cite{tolias2013aggregate-f56} retrieval, and is the system we extend in this work. Scene-prior systems include VGGT-SLAM~\cite{maggio2025vggt-slam:-22f}, which incrementally aligns VGGT submaps, and SLAM3R~\cite{liu2025slam3r:-64a}, which fuses local reconstructions into a global scene. However, to the best of our knowledge, no reconstruction-prior-based SLAM works have investigated how depth might be used as an input.

\section{Methods}\label{sec:methods}

Fig.~\ref{fig:pipeline} displays the method overview, highlighting how Pow3R-SLAM works, and where both depth and the ICP hybrid approach play a role in the system.

\subsection{Preliminaries}\label{sec:prelims}

Throughout the text, we adopt the conventions of MASt3R-SLAM~\cite{murai2025mast3r-slam:-690}. Given two views $i, j$, Pow3R~\cite{jang2025pow3r:-27b} defines a network $\mathcal{F}$ that takes an image pair $\mathcal{I}^i, \mathcal{I}^j \in \mathbb{R}^{H\times W\times3}$ and outputs pointmaps $\mathbf{X}^i_i, \mathbf{X}^j_i, \mathbf{X}^j_j$ (where $\mathbf{X}^j_i$ denotes the pointmap of image $\mathcal{I}^j$ in the coordinate system of the camera from view $i$) and their associated confidences $\mathbf{C}^i_i, \mathbf{C}^j_i, \mathbf{C}^j_j$. Pow3R can optionally take a subset of auxiliary information $\Omega \subseteq \{\mathbf{Z}^i, \mathbf{Z}^j, \mathbf{K}^i, \mathbf{K}^j, \mathbf{P}_{ij}\}$, including depthmaps $\mathbf{Z}^i, \mathbf{Z}^j \in \mathbb{R}^{H\times W\times 1}$ (called $\mathbf{D}$ in~\cite{jang2025pow3r:-27b}) with their validity masks $\mathbf{M}^i, \mathbf{M}^j \in \{0, 1\}^{H\times W}$, which accompany a depth map rather than being a prior of their own and indicate pixels with valid depth, camera intrinsics $\mathbf{K}^i, \mathbf{K}^j\in \mathbb{R}^{3\times3}$, and relative pose $\mathbf{P}_{ij} \in \mathbb{R}^{4\times4}$. We write a forward pass with priors $\Omega$ as $\mathcal{F}_P(\mathcal{I}^i, \mathcal{I}^j, \Omega)$, outputting $(\mathbf{X}^i_i, \mathbf{C}^i_i), (\mathbf{X}^j_i, \mathbf{C}^j_i), (\mathbf{X}^j_j, \mathbf{C}^j_j)$. As Pow3R-SLAM is an RGB-D approach, we assume that $\mathbf{Z}^i, \mathbf{Z}^j$ are always available. $\mathbf{P}_{ij}$ is what we solve for and is never supplied, so $\Omega = (\mathbf{Z}^i, \mathbf{Z}^j, \mathbf{K})$ throughout this work. Pow3R-SLAM's calibrated mode assumes that $\mathbf{K}^i = \mathbf{K}^j$ for all frames (as does MASt3R-SLAM's), and a fixed calibration is required regardless to register depth to RGB. All results use this mode.

We maintain the same pose formulation as MASt3R-SLAM, with $\mathbf{T} \in \mathbf{Sim}(3)$, and updates $\tau \in \mathfrak{sim}3$ applied by a left-plus operator. $\mathbf{T} \leftarrow \tau \oplus \mathbf{T} \triangleq \mathrm{Exp}(\tau) \circ \mathbf{T}$, where $\mathbf{T}$ holds a rotation $\mathbf{R} \in \mathbf{SO}(3)$, a translation $\mathbf{t} \in \mathbb{R}^3$, and a scale $s > 0$.

Pow3R-SLAM uses a known pinhole camera as its primary setting. Images with lens distortion are undistorted first. $\mathbf{K}$ is adjusted when the image is resized for Pow3R (long side at 512 pixels) and this resized $\mathbf{K}$ is used throughout the pipeline. Let $\mathcal{R}_{\mathbf{K}}(\mathbf{X})[\mathbf{p}] \triangleq z(\mathbf{X}[\mathbf{p}])\,
\mathbf{K}^{-1}[u,v,1]^\top$ be the operation that keeps a pointmap's depth $z(\cdot)$ at pixel
$\mathbf{p} = (u,v)$ and returns the point to its calibrated ray. Both self-view pointmaps are
passed through $\mathcal{R}_{\mathbf{K}}$ before use, so only the network's depths reach
the estimator.

\subsection{Known Depth and Calibration}\label{sec:depth_cal}
A primary focus of this work is the integration of depth data into the pipeline. Sensor depth is involved in three areas of Pow3R-SLAM (Fig.~\ref{fig:pipeline}), denoted: \textbf{cond}, \textbf{scale}, and \textbf{anchor}.

\subsubsection{\textbf{cond}} (conditioning) feeds depth in as an optional input to Pow3R. Following Pow3R's convention~\cite{jang2025pow3r:-27b}, each depth map is normalized by its
mean over valid pixels, $\mathbf{Z}/\bar{z}$ with $\bar{z} = \text{avg}_{\mathbf{M}}(\mathbf{Z})$,
together with its validity mask $\mathbf{M} = [\mathbf{Z} > 0]$. The intrinsics are not passed as
a matrix, and are instead expanded into a dense ray image, evaluating $\mathbf{K}^{-1}[u, v, 1]^\top$ at
every pixel $(u, v)$. Both maps are then cut into the same $16\times16$ patch grid as the image,
and embedded into one token per patch. Those tokens are injected inside the network's first encoder block, where a learned projection of each is added to the token of the image patch it covers. Pow3R therefore receives $\mathcal{F}_P(\mathcal{I}^i, \mathcal{I}^j, \Omega)$ with
$\Omega = (\mathbf{Z}^i, \mathbf{Z}^j, \mathbf{K})$.

\subsubsection{\textbf{scale}} restores metric scale from the (metric) depth images. In each
forward pass, one scalar is estimated from the self-view pointmap $\mathbf{X} = \mathbf{X}^i_i$
against its sensor depth $\mathbf{Z} = \mathbf{Z}^i$, and applied to both
outputs of that pass ($\mathbf{X}^i_i$ and $\mathbf{X}^j_i$, which share a single gauge, i.e.\ the same unknown scale and coordinate frame). Let the per-pixel depth ratio $r_\mathbf{p} = \frac{\mathbf{Z}[\mathbf{p}]}{z(\mathbf{X}[\mathbf{p}])}$, over valid pixels $\mathbf{p}$ with $\mathbf{M}[\mathbf{p}] = 1$ and $z(\mathbf{X}[\mathbf{p}]) > 0$ (such that $z(\mathbf{x})$ is the depth of a point $\mathbf{x} \in \mathbf{X}$). Then, given $\text{median}(r_\mathbf{p}) = \bar{r}$, our scaling formula is defined as:

\begin{equation}
    \label{eq:scaling}
    s(\mathbf{X}, \mathbf{Z}) \triangleq \exp(\text{avg}_\mathbf{p}(\log(\text{clip}(r_\mathbf{p}, 0.1 \cdot \bar{r}, 10 \cdot\bar{r})))),
\end{equation}

where $\text{clip}(a, b, c) = \min(\max(a, b), c)$, applied when at least $50$ pixels are valid (a floor against near-empty depth maps below which the pass is left unscaled). The geometric mean is the minimizer of the squared log-depth residuals that our solvers use (the log-depth row of Eq.~\ref{eq:corr}, and Eq.~\ref{eq:err_back_depth}), and the clip rejects outliers. The scale is the geometric mean of the per-pixel depth ratios, with ratios more than $10\times$ from their median pulled back to that bound.

\subsubsection{\textbf{anchor}} brings the sensor into the backend. Prior to each backend solve, the depth of each keyframe pointmap is replaced along its known rays by the sensor's depth wherever the sensor has data. The substituted depths are not clamped in the solution, acting instead as a soft constraint in the backend optimization (Sec.~\ref{sec:backend}).

\subsection{Pointmap Matching and Fusion}\label{sec:matching}

We adopt the projective matching approach of MASt3R-SLAM. For a function $\psi(\mathbf{x})$ that normalizes a point to a unit ray, the match for $\mathbf{x} \in \mathbf{X}^j_i$ is, following~\cite{murai2025mast3r-slam:-690}, the pixel $\mathbf{p}^* = \argmin_{\mathbf{p}} \| \psi([\mathbf{X}^i_i]_{\mathbf{p}}) - \psi(\mathbf{x}) \|^2$, solved by a per-point Levenberg--Marquardt with ten iterations. For new keyframes we initialize from an identity mapping, we employ next-frame initialization for tracking, accept a match when its ray residual converges below $10^{-6}$, and reject outliers at $>0.1$\,m in 3D space. The only difference is that we do not use per-pixel features to further refine the matches, as Pow3R has no such features.

After each tracked frame (Sec.~\ref{sec:tracking}), we use the same criteria as MASt3R-SLAM to add new keyframes. Let $\omega_k$ be the smaller of the fraction of keyframe pixels matched and the fraction of distinct keyframe pixels those matches land on. A keyframe is added when $\omega_k < 0.333$. Once pose is solved, we update the keyframe's geometry via $\mathbf{X}^k_k \leftarrow \mathbf{T}_{kf} \mathbf{X}^k_f$ (where $\mathbf{X}^k_k$ is the keyframe pointmap, $\mathbf{T}_{kf}$ is the relative pose between the incoming frame and the keyframe (Sec.~\ref{sec:tracking}), and $\mathbf{X}^k_f$ the keyframe's pointmap in the incoming frame's coordinates). While MASt3R-SLAM uses a running weighted average filter, for each pixel we instead keep the most confident prediction the keyframe has received:

\begin{equation}
    \label{eq:pointmap_fuse}
    (\Tilde{\mathbf{X}}^k_k[\mathbf{p}], \Tilde{\mathbf{C}}^k_k[\mathbf{p}])
    \leftarrow (\mathbf{X}^k_k[\mathbf{p}], \mathbf{C}^k_f[\mathbf{p}]) \ \text{if} \
    \mathbf{C}^k_f[\mathbf{p}] > \Tilde{\mathbf{C}}^k_k[\mathbf{p}] .
\end{equation}

Since each pass has its own scalar (from Eq.~\ref{eq:scaling}), averaging predictions would average scale error into the keyframe.

\subsection{Tracking}\label{sec:tracking}

Tracking estimates the pose of each incoming frame $f$ relative to the current keyframe $k$ from a single pass $\mathcal{F}_{P}(\mathcal{I}^f, \mathcal{I}^k, \Omega)$ per frame, following MASt3R-SLAM's frame-to-keyframe scheme. For an incoming point transported into the keyframe camera $\mathbf{x}_m$, and the keyframe's own canonical point at the matched pixel $\mathbf{y}_n$, let $\mathbf{x}_m = \mathbf{T}_{kf}\mathcal{R}_\mathbf{K}(\mathbf{X}^f_f)_m$ and $\mathbf{y}_n = \mathcal{R}_\mathbf{K}(\Tilde{\mathbf{X}}^k_k)_n$ (Sec.~\ref{sec:prelims}).

With $\Pi_\mathbf{K}$ the projection under $\mathbf{K}$, a correspondence $(m, n)$ in the match set $\mathbf{m}_{f, k}$ gives:

\begin{equation}
    \label{eq:corr}
    \mathbf{r}_{mn} =
    \begin{bmatrix}
        \Pi_\mathbf{K}(\mathbf{y}_n) - \Pi_\mathbf{K}(\mathbf{x}_m) \\
        \log z(\mathbf{y}_n) - \log z(\mathbf{x}_m)
    \end{bmatrix}.
\end{equation}

Following the weighting of Eq. (5) in~\cite{murai2025mast3r-slam:-690}, a residual with confidence $q$ carries the covariance $w(q,\sigma^2) = \sigma^2/q$ when $q>q_{\min}$ and is discarded ($w=\infty$) otherwise. $\lVert\mathbf{r}\rVert_{\rho,w}$ denotes the Huber norm (threshold 1.345) of the whitened residual $\mathbf{r}/\sqrt{w}$.
In tracking, Pow3R provides no per-match confidence, so $q \equiv 1$, and outliers are rejected by the matching tests of Sec.~\ref{sec:matching} (the ray-residual convergence check and the $0.1$\,m 3D rejection) rather than by the confidence gate.

Tracking minimizes the reprojection error:
\begin{equation}
    \label{eq:err_repro}
    E_{\Pi} = \sum_{(m,n)\in\mathbf{m}_{f,k}}
    \left\lVert \Pi_\mathbf{K}(\mathbf{y}_n) -
    \Pi_\mathbf{K}(\mathbf{x}_m)\right\rVert_{\rho,\,w(1, \sigma_{\text{px}}^2)} ,
\end{equation}
with $\sigma_{\text{px}} = 1$ pixel. We use Gauss--Newton for at most 50 iterations to optimize reprojection error and recover the relative pose $\mathbf{T}_{kf}$. As in MASt3R-SLAM, a second, lightly weighted term enters the same solve (the log-depth row of Eq.~\ref{eq:corr}). This prevents degeneracy under pure rotation, where reprojection leaves depth unobserved, and it alone observes the scale of $\mathbf{T}_{kf} \in \textbf{Sim}(3)$, since a similarity scaling leaves $\Pi_{\mathbf{K}}$ invariant while $\log z$ shifts by $\log s$. We use a loose $\sigma^t_z = 10$ (superscripts $t$ and $g$ denoting tracking and global solves), admitting a small percent of per-frame scale noise for the backend to reconcile. If fewer than $5\%$ of pixels survive the matching, or the normal equations fail to factorize, the frame is deemed lost and relocalization (Sec.~\ref{sec:relocalization}) is triggered. Upon adding a new keyframe, a bidirectional edge is added to the previous keyframe.

\subsection{Loop Closure, Relocalization, and Backend}

\subsubsection{Loop Closure} recognizes re-visited areas and corrects the relative keyframe poses. We use the same ASMK~\cite{tolias2013aggregate-f56, tolias2020learning-198} retrieval as MASt3R-SLAM, but in place of
MASt3R's encoder tokens we use an unconditioned Pow3R encoder pass over the
keyframe image, computed once per keyframe. We refit the training-free ASMK head to Pow3R's tokens, re-estimating the principal component analysis (PCA) whitening and reducing the $k$-means codebook from 65536 to 16384 centroids. New keyframes query the top 3 candidates above a score threshold of 0.005. The candidates are each decoded by a Pow3R pass in each direction and matched projectively as in Sec.~\ref{sec:matching}, with a loop edge added when the minimum match fraction is at least $\omega_l = 0.25$ (the edge to the preceding keyframe is unconditionally added).

\subsubsection{Relocalization}\label{sec:relocalization} occurs when tracking is lost (Sec.~\ref{sec:tracking}). We follow the same approach as MASt3R-SLAM here, querying the retrieval database with the same score threshold but a stricter acceptance rule. The retrieved frame is added as a keyframe only if every retrieved candidate clears a match-fraction floor of $0.3$, after which its pose is initialized from the first candidate and a backend solve is run.

\subsubsection{Backend}\label{sec:backend} We reuse the second-order optimization scheme of MASt3R-SLAM, bidirectionally minimizing the calibrated residual over all edges via Gauss--Newton with a sparse Cholesky solve, with the first keyframe's pose held fixed, and at most ten iterations per new keyframe.
There are two primary differences in our method.

First, prior to each solve, an \textbf{anchor} step replaces (only for the backend solve) the depth for each keyframe pointmap with its sensor values along the rays that form the pointmap, where the sensor has data ($\hat{\mathbf{X}}^i_i = \mathcal{R}_\mathbf{K}(\Tilde{\mathbf{X}}^i_i)$ with $z \leftarrow \mathbf{Z}^i$ on $\mathbf{M}^i$).

As the sensor is typically more accurate where it has data, we set $\sigma^g_z = 0.1$ for the log-depth residual here (against $\sigma^t_z=10$ in tracking).

For each edge $(i, j)$ in the set of all edges $\mathcal{E}$, traversed in both directions $\mathcal{E}^{\pm}$, with $\mathbf{T}_{ij} = \mathbf{T}^{-1}_{WC_i}\mathbf{T}_{WC_j}$, $\hat{\mathbf{x}}_m = \mathbf{T}_{ij}\,\mathcal{R}_{\mathbf{K}}\bigl(\hat{\mathbf{X}}^j_j\bigr)_m$, and
$\hat{\mathbf{y}}_n = \mathcal{R}_{\mathbf{K}}\bigl(\hat{\mathbf{X}}^i_i\bigr)_n$, the backend minimizes the sum $E^g_\Pi+E^g_z$ of:
\begin{equation}
    \label{eq:err_back_proj}
    E^g_{\Pi} = \sum_{(i,j)\in\mathcal{E}^{\pm}} \sum_{(m,n)\in\mathbf{m}_{ij}}
    \left\lVert \Pi_\mathbf{K}(\hat{\mathbf{y}}_n) - \Pi_\mathbf{K}(\hat{\mathbf{x}}_m)\right\rVert_{\rho,\,w(w_{mn}, \sigma_{\text{px}}^2)} ,
\end{equation}

\begin{equation}
    \label{eq:err_back_depth}
    E^g_{z} = \sum_{(i,j)\in\mathcal{E}^{\pm}} \sum_{(m,n)\in\mathbf{m}_{ij}}
    \left\lVert \log z(\hat{\mathbf{y}}_n) - \log z(\hat{\mathbf{x}}_m)\right\rVert_{\rho,\,w(w_{mn}, (\sigma_z^{\,g})^2)} ,
\end{equation}

with hats indicating an anchored pointmap, $\sigma_{\text{pix}} = 1$ px as in tracking, and $w_{mn}$ (Eq.~\ref{eq:confs}) filling the role of $q_{mn}$.

Anchoring is disabled when scale is disabled, and the anchored copy never leaves the backend.

Second, as we do not have MASt3R's matching confidence, we rebuild the per-correspondence weight from Pow3R's confidences as:

\begin{equation}
    \label{eq:confs}
    w_{mn} = \mathbf{Q}_{mn}\Tilde{\mathbf{C}}^i_{i, m}\Tilde{\mathbf{C}}^j_{j, n}, \ \text{where} \  \mathbf{Q}_{mn} = \phi(\sqrt{\mathbf{C}^i_{i, m}\mathbf{C}^j_{i, n}}),
\end{equation}

where $\phi$ is a piecewise-linear quantile map with 1024 knots, fitted once so that its output distribution matches MASt3R-SLAM's matching confidence over 109630 correspondences from eight development sequences (Sec.~\ref{sec:results}).
$\textbf{Q}_{mn}$ reproduces MASt3R-SLAM's gate $\mathbf{Q}_{mn} > 1.5$ without threshold re-tuning or retraining. The two $\Tilde{C}$ factors down-weight low-confidence geometry, and in practice the
$\mathbf{Q}_{mn} > 1.5$ gate carries most of the effect of $w_{mn}$.

\subsection{Map-Only Keyframes}\label{sec:maponlykf}

The keyframe rate that is ideal for pose estimation leads to maps that are too sparse, but raising it changes the estimator, so we look to decouple map density from the estimator. For this, we add map-only keyframes, which carry geometry but no estimator state. A network-tracked frame with $0.333 \leq \omega_k < 0.45$ is stored when at least 5 input frames have passed since the previous addition, up to 200 per sequence. For each map-only keyframe $f$, we store its rescaled pointmap $\mathbf{X}^f_f$, confidence $\mathbf{C}^f_f$, image $\mathcal{I}^f$, and its pose relative to its keyframe $\mathbf{T}_{kf}$.

At export, the world pose is recomposed against the keyframe's final pose, $\mathbf{T}_{{WC}_{f}} = \mathbf{T}_{WC_k}\mathbf{T}_{kf}$ in $\mathbf{Sim}(3)$, so backend corrections and loop closures propagate to them. The final exported map is the union of the keyframe and map-only pointmaps under a single confidence threshold. This densifies the map, without modifying trajectories.

\subsection{Hybrid ICP Variant}\label{sec:icp}

In both MASt3R- and Pow3R-SLAM, the network forward pass dominates the tracker's cost, which greatly affects overall runtime. Inspired by MASt3R-SLAM's success when simulating real-time performance by skipping every 2nd frame and existing depth-based approaches, our \textbf{hybrid} variant processes every 2nd frame via point-to-plane ICP~\cite{chen1992object-868, rusinkiewicz2001efficient-b64} on the depth sensor input, replacing the network call for those frames. Let $\mathbf{x}_{\mathbf{p}} = \mathbf{Z}^f[\mathbf{p}]\,\mathbf{K}^{-1}[u,v,1]^\top$ be frame $f$'s sensor depth back-projected along its rays and $\mathbf{q}_{\mathbf{p}} = \Pi_\mathbf{K}(\mathbf{T}_{kf}\,
\mathbf{x}_{\mathbf{p}})$ its projective association (nearest pixel) into the keyframe, recomputed at every ICP iteration. The sum runs over the set $\mathbf{P}$ of pixels at which both the sensor depth and the keyframe pointmap carry a valid central-difference normal (excluding borders, hole neighbours, and depth steps above $5\%$). With $\mathbf{n_{q_{p}}}$ the keyframe's normal at $\mathbf{q_p}$, the point-to-plane error is minimized against the keyframe's canonical pointmap~\cite{newcombe2011kinectfusion:-793, chen1992object-868}:

\begin{equation}
    \label{eq:icp}
    E_{\text{icp}} = \sum_{\mathbf{p} \in \mathbf{P}}
    \lVert( \mathbf{n}_{\mathbf{q_p}}^{\top}
    \bigl( \mathbf{T}_{kf}\,\mathbf{x}_{\mathbf{p}} -
    \tilde{\mathbf{X}}^k_k[\mathbf{q}_{\mathbf{p}}] \bigr)) \rVert_{\rho'},
\end{equation}

where $\rho'$ is a Huber kernel at 2 cm (a metric distance, unlike $\rho$). Our solve is coarse-to-fine over 3 levels with iteration counts $(6, 4, 3)$ on $\mathbf{SE}(3)$ since the frame inherits its keyframe's scale, gated on correspondence distance $(10 \text{cm})$ and normal rotation $(30^\circ)$, and falls back to network tracking whenever the pose is not finite, the finest-level cost rises by $>5\%$, or $<30\%$ of the associable pixels are inliers.

ICP-tracked frames never become keyframes, reach the backend, or add points to the map.

\section{Results}\label{sec:results}

We evaluate localization on TUM RGB-D~\cite{sturm2012benchmark-29f}, 7-Scenes~\cite{glocker2013real-time-7bb}, Replica (synthetic)~\cite{straub2019replica-37a}, and ETH3D-SLAM~\cite{schöps2019bad-8c9}, and geometry on 7-Scenes and Replica, which have ground-truth geometry.

7-Scenes provides its Kinect colour and depth images unregistered, with only the depth camera's intrinsics, which misplaces the edges between depth and colour pixels by 20--40 px. For every system that uses depth, we therefore resample each colour image into the depth camera through the depth image. To calibrate, we take the colour focal length and radial distortion (from COLMAP~\cite{schönberger2016structure-from-motion-314} self-calibration of the colour images) and the 2.7 cm baseline and principal point from depth-colour edge alignment. One calibration is shared by all scenes, and neither step uses ground truth or trajectory error. The fitting corpus of $\phi$ (Sec.~\ref{sec:backend}) overlaps the evaluation panel on four sequences (\emph{fr1/teddy}, \emph{room0}, \emph{pumpkin}, \emph{stairs}). As $\phi$ only preserves MASt3R-SLAM's threshold meaning, and refitting it without the four moves them by at most $18$ mm (the largest, \emph{stairs}, an improvement), this is not a material leak.
The retrieval codebook was fitted on five TUM fr2/fr3 and eight ETH3D-SLAM training sequences, which we exclude from evaluation.

All runs use a Ryzen 5900X 3.7\,GHz CPU and an NVIDIA GeForce RTX\,4090. As in MASt3R-SLAM's protocol, every 2nd frame is processed. Pow3R-SLAM and MASt3R-SLAM accuracy is deterministic, and every runtime is the mean of two exclusive repeats, with the half-range reported in Table~\ref{tab:ate2}. All results use the single-threaded mode, in which the front end waits for the backend to drain. Using either system's non-blocking mode changes wall time by under $5\%$.

\subsection{Pose Estimation}\label{sec:pose_estimation}

Following the convention of MASt3R-SLAM, we report the root-mean-square-error (RMSE) of the absolute trajectory error (ATE) in meters (Table~\ref{tab:ate1}). Pose estimation results are from the Sim(3)-aligned keyframe trajectory. We re-run MASt3R-SLAM for direct comparison, which reproduces the published values of~\cite{murai2025mast3r-slam:-690} to the reported precision on all 16
sequences from TUM and 7-Scenes on native images. On 7-Scenes, all tables report it on the registered images. We also re-run ORB-SLAM3~\cite{campos2021orb-slam3:-eb8} in RGB-D mode with calibration, at a median of five runs unless otherwise noted (v1.0, loop closure on, stock ORB parameters). In our tables the best result is \textbf{bolded}, the second best \underline{underlined}, and arrows give the better direction.

\begin{table}[t]
\centering
\caption{\textbf{ATE root-mean square error (RMSE) (m) on TUM RGB-D fr1 and 7-Scenes}. \textbf{Bold}, \underline{underline}: best, second best. $\dagger$: calibrated monocular result as reported in~\cite{murai2025mast3r-slam:-690}. Fresh: our re-run. X: lost track, with means over completed scenes. $\ast$: unscaled SE(3) alignment.}
\label{tab:ate1}
{\footnotesize\setlength{\tabcolsep}{3pt}\renewcommand{\arraystretch}{0.95}\begin{tabular}{lrrrrrrr}
    \toprule
     &  & \multicolumn{2}{c}{\scriptsize\textbf{ORB-SLAM3}} &  & \multicolumn{3}{c}{\scriptsize\textbf{\mbox{Pow3R-SLAM}}} \\
    \cmidrule(lr){3-4}\cmidrule(lr){6-8}
    \scriptsize\textbf{scene} & \scriptsize\textbf{DROID\textsuperscript{\dag}\,$\downarrow$} & \scriptsize\textbf{Sim3\,$\downarrow$} & \scriptsize\textbf{SE3\textsuperscript{*}\,$\downarrow$} & \scriptsize\textbf{\shortstack[c]{MASt3R\\fresh\,$\downarrow$}} & \scriptsize\textbf{Sim3\,$\downarrow$} & \scriptsize\textbf{SE3\textsuperscript{*}\,$\downarrow$} & \scriptsize\textbf{\shortstack[c]{hyb.\\Sim3\,$\downarrow$}} \\
    \midrule
    \multicolumn{8}{l}{\scriptsize\textbf{TUM RGB-D fr1}} \\
    \scriptsize 360 & \scriptsize 0.111 & \scriptsize 0.134 & \scriptsize 0.228 & \scriptsize 0.049 & \scriptsize \textbf{0.040} & \scriptsize 0.048 & \scriptsize \underline{0.042} \\
    \scriptsize desk & \scriptsize 0.018 & \scriptsize 0.018 & \scriptsize 0.018 & \scriptsize \textbf{0.016} & \scriptsize 0.019 & \scriptsize 0.035 & \scriptsize \underline{0.017} \\
    \scriptsize desk2 & \scriptsize 0.042 & \scriptsize X & \scriptsize X & \scriptsize \underline{0.024} & \scriptsize \underline{0.024} & \scriptsize 0.093 & \scriptsize \textbf{0.022} \\
    \scriptsize floor & \scriptsize \textbf{0.021} & \scriptsize X & \scriptsize X & \scriptsize \underline{0.025} & \scriptsize \textbf{0.021} & \scriptsize 0.025 & \scriptsize \underline{0.025} \\
    \scriptsize plant & \scriptsize \textbf{0.016} & \scriptsize 0.022 & \scriptsize 0.026 & \scriptsize 0.020 & \scriptsize \underline{0.019} & \scriptsize 0.020 & \scriptsize \textbf{0.016} \\
    \scriptsize room & \scriptsize 0.049 & \scriptsize 0.073 & \scriptsize 0.084 & \scriptsize 0.061 & \scriptsize \textbf{0.042} & \scriptsize 0.044 & \scriptsize \underline{0.046} \\
    \scriptsize rpy & \scriptsize 0.026 & \scriptsize 0.035 & \scriptsize 0.040 & \scriptsize 0.027 & \scriptsize \underline{0.019} & \scriptsize 0.023 & \scriptsize \textbf{0.017} \\
    \scriptsize teddy & \scriptsize 0.048 & \scriptsize X & \scriptsize X & \scriptsize \underline{0.041} & \scriptsize 0.049 & \scriptsize 0.066 & \scriptsize \textbf{0.032} \\
    \scriptsize xyz & \scriptsize 0.012 & \scriptsize 0.012 & \scriptsize 0.013 & \scriptsize 0.009 & \scriptsize \underline{0.006} & \scriptsize 0.007 & \scriptsize \textbf{0.005} \\
    \scriptsize \textbf{mean} & \scriptsize 0.038 & \scriptsize 0.049 (6) & \scriptsize 0.068 (6) & \scriptsize 0.030 & \scriptsize \underline{0.027} & \scriptsize 0.040 & \scriptsize \textbf{0.025} \\
    \midrule
    \multicolumn{8}{l}{\scriptsize\textbf{7-Scenes}} \\
    \scriptsize chess & \scriptsize 0.036 & \scriptsize \underline{0.033} & \scriptsize 0.038 & \scriptsize 0.038 & \scriptsize \textbf{0.027} & \scriptsize 0.035 & \scriptsize \textbf{0.027} \\
    \scriptsize fire & \scriptsize 0.027 & \scriptsize \textbf{0.023} & \scriptsize 0.028 & \scriptsize 0.029 & \scriptsize 0.025 & \scriptsize 0.031 & \scriptsize \underline{0.024} \\
    \scriptsize heads & \scriptsize 0.025 & \scriptsize \underline{0.014} & \scriptsize 0.017 & \scriptsize 0.017 & \scriptsize \textbf{0.012} & \scriptsize 0.015 & \scriptsize \underline{0.014} \\
    \scriptsize office & \scriptsize \textbf{0.066} & \scriptsize 0.096 & \scriptsize 0.098 & \scriptsize 0.092 & \scriptsize \underline{0.085} & \scriptsize 0.085 & \scriptsize 0.086 \\
    \scriptsize pumpkin & \scriptsize 0.127 & \scriptsize 0.127 & \scriptsize 0.127 & \scriptsize 0.084 & \scriptsize \textbf{0.077} & \scriptsize 0.087 & \scriptsize \underline{0.083} \\
    \scriptsize kitchen & \scriptsize 0.040 & \scriptsize 0.047 & \scriptsize 0.048 & \scriptsize 0.060 & \scriptsize \textbf{0.038} & \scriptsize 0.038 & \scriptsize \underline{0.039} \\
    \scriptsize stairs & \scriptsize \underline{0.026} & \scriptsize 0.045 & \scriptsize 0.049 & \scriptsize \textbf{0.016} & \scriptsize 0.044 & \scriptsize 0.051 & \scriptsize 0.062 \\
    \scriptsize \textbf{mean} & \scriptsize 0.049 & \scriptsize 0.055 & \scriptsize 0.058 & \scriptsize \underline{0.048} & \scriptsize \textbf{0.044} & \scriptsize 0.049 & \scriptsize \underline{0.048} \\
    \bottomrule
  \end{tabular}
}
\end{table}

\subsubsection{TUM RGB-D}\label{sec:traj_tum} Pow3R-SLAM has the lowest mean trajectory error on the TUM sequences (Table~\ref{tab:ate1}), in both hybrid and regular modes. ORB-SLAM3 loses track on $3/9$ sequences (\emph{desk2}, \emph{floor}, \emph{teddy}), while all other systems complete all 9. Pow3R-SLAM improves on MASt3R-SLAM on $6/9$ sequences, and on the mean ($0.027$ vs.\ $0.030$ m), with the largest relative gains on \emph{room} and, in hybrid mode, \emph{teddy}.

\subsubsection{7-Scenes}\label{sec:7-scenes}

With colour registered to depth, Pow3R-SLAM has the lowest mean on 7-Scenes (0.044\,m, against 0.048\,m for MASt3R-SLAM and 0.055\,m for ORB-SLAM3), better than MASt3R-SLAM on $6/7$ sequences. \emph{stairs} is the exception (0.044 vs.\ 0.016\,m), which we analyze in Sec.~\ref{sec:limitations} and Fig.~\ref{fig:stairs}. The registration lowered the 7-Scenes mean absolute trajectory error (ATE) of Pow3R-SLAM from 0.065 to 0.044\,m and of ORB-SLAM3 from 0.059 to 0.055\,m. MASt3R-SLAM receives the same registered images, although it uses no depth. On native images, its 7-Scenes ATE was 0.047\,m, which became 0.048\,m.

\subsubsection{Replica}\label{sec:traj_replica}

On Replica, Pow3R-SLAM significantly outperforms MASt3R-SLAM (Table~\ref{tab:ate2}). Mean ATE is almost half that of MASt3R-SLAM ($0.009$ vs. $0.015$ m), with tighter trajectories on $5/8$ sequences. Replica's rendered depth is complete and noise-free, which strengthens conditioning and scale. ORB-SLAM3 is the most accurate system here (mean $0.005$ m), for which the classical tracking benefits from exact, hole-free depth, and richly textured renders.

\subsubsection{ETH3D-SLAM}\label{sec:traj_eth3d} Due to ETH3D-SLAM's fast camera motion, we use all frames, as MASt3R-SLAM does. From its 61 training sequences we evaluate a 16-sequence panel which was fixed before any runs. Table~\ref{tab:ate2} reports the mean over completed sequences, the median, and the area under the success-versus-threshold curve (AUC) for thresholds up to $0.5$\,m (max.\ 8 on this panel, which is not comparable to the 61-sequence AUC of~\cite{murai2025mast3r-slam:-690}). While MASt3R-SLAM completes all 16, and Pow3R-SLAM does not complete 2 (\emph{planar\_3} spans $5\times$ the ground truth extent and \emph{repetitive} does not finish), Pow3R-SLAM's median is lower (0.0135 vs.\ 0.0280 m) and it wins $12/14$ sequences that both complete. Its mean over those 14 (0.059 m) is inflated by \emph{sofa\_2} at 0.66 m. A sequence counts as completed, for every system, when all keyframe poses are finite and the trajectory span is within $2\times$ the ground-truth span. In comparison, ORB-SLAM3 fails 6 scenes in ETH3D-SLAM, and of the 9 sequences all three algorithms complete its ATE lies between Pow3R-SLAM and MASt3R-SLAM (0.026 vs. 0.014 and 0.038 m, respectively). Counting all scenes, its AUC is 4.76, against 6.33 for Pow3R-SLAM and 7.36 for MASt3R-SLAM, which completes every sequence. The hybrid mode completes the same 14 scenes at a lower mean than MASt3R-SLAM (0.034 vs. 0.039 m on those 14, better on 12/14).

\subsubsection{Per-frame and Unscaled Scoring} Over the 24-scene panel, the keyframe ATE is a like-for-like comparison between MASt3R- and Pow3R-SLAM, which insert 537 and 525 keyframes, respectively (per-sequence counts differ by up to 8), with hybrid inserting 503. Scored on every frame under rigid \textbf{SE}(3) with no scale fitted, Pow3R-SLAM's ATE is 0.037\,m against MASt3R-SLAM's 0.116\,m, which is $3.1\times$ lower. As MASt3R-SLAM has no metric scale, this gap measures the value of the scale site rather than of the prediction prior (Table~\ref{tab:ablation}, L3 vs. L4). Scoring all frames flips only the hybrid's $\mathbf{Sim}(3)$, where its mean rises
from 0.026 to 0.035\,m against MASt3R-SLAM's 0.030 to 0.034\,m. Under $\mathbf{SE}(3)$ the hybrid
stays $2.9\times$ ahead, and per-frame loss is likely because ICP tracks worse with sparse depth.

\begin{table}[t]
\centering
\caption{\textbf{Replica, ETH3D-SLAM and runtime.} Keyframe ATE RMSE (m), Sim(3), with the
unscaled SE(3) mean. ETH3D-SLAM: over each system's completed sequences (count in brackets).
Runtime: wall time summed over the 24-scene panel (TUM, 7-Scenes, Replica), at full and subsample-2 (s2). Mean of 2 exclusive repeats per system ($\pm$ half-range), with MASt3R-SLAM timed on native images. FPS: frames per second at s2 of loop time excluding model load.}
\label{tab:ate2}
{\footnotesize\setlength{\tabcolsep}{3pt}\renewcommand{\arraystretch}{0.95}\begin{tabular}{lrrr}
    \toprule
    \scriptsize \textbf{scene / metric} & \scriptsize \textbf{\shortstack[c]{MASt3R\\fresh}} & \scriptsize \textbf{\shortstack[c]{Pow3R-\\SLAM}} & \scriptsize \textbf{\shortstack[c]{Pow3R-\\SLAM hyb.}} \\
    \midrule
    \scriptsize office0 & \scriptsize 0.013 & \scriptsize \underline{0.009} & \scriptsize \textbf{0.008} \\
    \scriptsize office1 & \scriptsize \textbf{0.007} & \scriptsize \underline{0.009} & \scriptsize 0.013 \\
    \scriptsize office2 & \scriptsize \underline{0.022} & \scriptsize \textbf{0.007} & \scriptsize \textbf{0.007} \\
    \scriptsize office3 & \scriptsize 0.027 & \scriptsize \textbf{0.010} & \scriptsize \underline{0.012} \\
    \scriptsize office4 & \scriptsize 0.021 & \scriptsize \textbf{0.009} & \scriptsize \underline{0.011} \\
    \scriptsize room0 & \scriptsize \textbf{0.006} & \scriptsize \underline{0.007} & \scriptsize \underline{0.007} \\
    \scriptsize room1 & \scriptsize 0.016 & \scriptsize \underline{0.007} & \scriptsize \textbf{0.006} \\
    \scriptsize room2 & \scriptsize \textbf{0.011} & \scriptsize \underline{0.013} & \scriptsize \underline{0.013} \\
    \scriptsize \textbf{mean, Sim(3)\,$\downarrow$} & \scriptsize 0.0153 & \scriptsize \textbf{0.0087} & \scriptsize \underline{0.0096} \\
    \scriptsize \textbf{mean, SE(3)\,$\downarrow$} & \scriptsize 0.0930 & \scriptsize 0.0107 & \scriptsize 0.0116 \\
    \midrule
    \scriptsize \textbf{ETH3D mean\,$\downarrow$} & \scriptsize 0.0400 (16) & \scriptsize 0.0588 (14) & \scriptsize 0.0343 (14) \\
    \scriptsize ETH3D median\,$\downarrow$ & \scriptsize 0.0280 & \scriptsize 0.0135 & \scriptsize 0.0104 \\
    \scriptsize ETH3D AUC\,$\uparrow$ & \scriptsize \textbf{7.3620} & \scriptsize 6.3345 & \scriptsize \underline{6.5190} \\
    \midrule
    \scriptsize \textbf{$\Sigma$ wall s2 (s)\,$\downarrow$} & \scriptsize 1443.3 $\pm$ 14.6 & \scriptsize \underline{902.6 $\pm$ 4.7} & \scriptsize \textbf{697.6 $\pm$ 1.0} \\
    \scriptsize $\Sigma$ wall full (s)\,$\downarrow$ & \scriptsize 2395.4 $\pm$ 0.6 & \scriptsize \underline{1555.1 $\pm$ 9.8} & \scriptsize \textbf{1146.5 $\pm$ 2.0} \\
    \scriptsize FPS mean (s2)\,$\uparrow$ & \scriptsize 14.0 & \scriptsize \underline{18.6} & \scriptsize \textbf{25.3} \\
    \bottomrule
  \end{tabular}
}
\end{table}

\subsection{Geometry Evaluation}\label{sec:geom_evaluation}

\begin{figure*}[ht]
\centering
\includegraphics[width=0.98\textwidth]{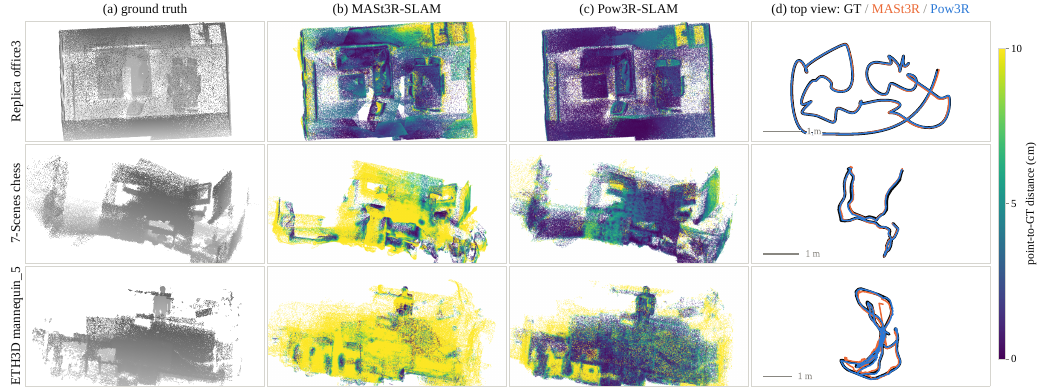}
\caption{\textbf{Qualitative reconstruction.} Rows: Replica \emph{office3}, 7-Scenes \emph{chess}, ETH3D-SLAM
\emph{mannequin\_5} (full frame rate). Columns: (a) reference cloud from ground-truth depth, (b) MASt3R-SLAM and (c) Pow3R-SLAM maps after alignment, coloured by distance to the reference (capped at 10\,cm), and (d) top view of
the per-frame trajectories (ground truth (GT) black, MASt3R-SLAM orange, Pow3R-SLAM blue).}
\label{fig:qual}
\end{figure*}

We evaluate the exported maps against reference clouds back-projected from every 20th ground-truth depth frame (culled at 4\,m, 8\,m on Replica), after \textbf{Sim}(3) alignment of the keyframe trajectory, with the estimate thresholded at $\mathbf{C} > 1.5$ and both clouds subsampled to 200k points. We report accuracy (mean estimate-to-reference distance), completion (the reverse), Chamfer (their average), as well as Chamfer RMSE. As noted in the supplementary materials of MASt3R-SLAM~\cite{murai2025mast3r-slam:-690}, mean Chamfer does not significantly penalize incorrect points, and as our maps cover more of the scene thanks to the map-only keyframes (11.9\,M vs.\ 2.7\,M points before subsampling, over the 15 map-scored scenes), RMSE Chamfer provides a fairer comparison. Table~\ref{tab:recon} gives the results over 7-Scenes and Replica between MASt3R- and Pow3R-SLAM. Pow3R-SLAM has better quality mapping versus MASt3R-SLAM on both 7-Scenes and (especially) Replica, with Chamfer RMSEs of 0.058 against 0.085 on 7-Scenes, and 0.027 against 0.056 on Replica, and both accuracy and completion improve on both datasets. Over the 15 scenes, mean Chamfer is 3.1 against 4.3\,cm (30\% lower), and RMSE Chamfer 4.1 against 6.9\,cm (40\% lower). Providing 7-Scenes' registered images to MASt3R-SLAM improved its 7-Scenes mean Chamfer from 6.7\,cm, to 5.5\,cm. Fig.~\ref{fig:qual} shows three scenes qualitatively.

\begin{figure}[ht]
\centering
\includegraphics[width=0.98\columnwidth]{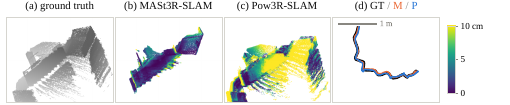}
\caption{\textbf{The failure case}. 7-Scenes \emph{stairs}, with the same columns as Fig.~\ref{fig:qual}, (d) GT / M(ASt3R) / P(ow3R).
MASt3R- vs.\ Pow3R-SLAM: 1.5 vs.\ 4.3\,M points, keyframe ATE 0.016 vs.\ 0.044\,m, accuracy 3.2 vs.\ 6.7\,cm, completion 10.5 vs.\ 5.9\,cm. Pow3R-SLAM's map is more complete but less accurate, which we attribute to the self-similar treads (Sec.~\ref{sec:limitations}).}
\label{fig:stairs}
\end{figure}

\begin{table}[ht]
\centering
\caption{\textbf{Reconstruction on 7-Scenes and Replica.} ATE for reference. Mean accuracy (Acc.), completion (Comp.), Chamfer (Cham.), and RMSE Chamfer, in m.}
\label{tab:recon}
{\scriptsize\setlength{\tabcolsep}{2pt}\renewcommand{\arraystretch}{0.95}\begin{tabular}{@{}l@{\;}lrrrrr@{}}
    \toprule
    \textbf{dataset} & \textbf{system} & \textbf{ATE\,$\downarrow$} & \textbf{Acc.\,$\downarrow$} & \textbf{Comp.\,$\downarrow$} & \textbf{Cham.\,$\downarrow$} & \textbf{Ch.\,RMSE\,$\downarrow$} \\
    \midrule
    7-Scenes & MASt3R-SLAM fresh & \underline{0.048} & 0.051 & \underline{0.058} & 0.055 & \underline{0.085} \\
     & Pow3R-SLAM & \textbf{0.044} & \textbf{0.048} & \textbf{0.038} & \textbf{0.043} & \textbf{0.058} \\
     & Pow3R-SLAM hyb. & \underline{0.048} & \underline{0.050} & \textbf{0.038} & \underline{0.044} & \textbf{0.058} \\
    \midrule
    Replica & MASt3R-SLAM fresh & 0.015 & 0.037 & 0.031 & 0.034 & 0.056 \\
     & Pow3R-SLAM & \textbf{0.009} & \textbf{0.022} & \textbf{0.018} & \textbf{0.020} & \textbf{0.027} \\
     & Pow3R-SLAM hyb. & \underline{0.010} & \underline{0.023} & \underline{0.019} & \underline{0.021} & \underline{0.028} \\
    \bottomrule
  \end{tabular}
}
\end{table}

\subsection{Runtime}

We also analyze runtime (bottom rows of Table~\ref{tab:ate2}). Compared to MASt3R-SLAM, which on the cross-dataset panel of all 24 scenes finishes in 1443 seconds with a mean FPS of 14.0, Pow3R-SLAM in its regular mode takes 903 seconds ($1.6\times$ faster), averaging 18.6 FPS, and its hybrid mode takes only 698 seconds ($2.1\times$ faster), averaging 25.3 FPS. At the full frame rate (full) the ratios are $1.5\times$ and $2.1\times$. The CPU-only ORB-SLAM3 (a sparse tracker with no network pass) takes 371 seconds, running at 51 FPS, for reference. On 7-Scenes, Pow3R-SLAM times include the colour registration.
Pow3R has no matching head, so MASt3R-SLAM's per-frame descriptor refinement is absent. The hybrid variant also skips the network pass on every 2nd frame, which occupies 2/3 of the tracker's time. Pow3R also has fewer parameters than MASt3R (556\,M vs.\ 695\,M), yet its network pass takes about as long (41 vs.\ 44\,ms per frame).

\subsection{Ablations}\label{sec:ablations}

To study the effects of each of Pow3R-SLAM's components, Table~\ref{tab:ablation} switches them on one at a time.
The single most impactful change is the introduction of priors to Pow3R, which takes the mean ATE from 0.083\,m (against MASt3R-SLAM's 0.030\,m) to 0.032\,m (L1 to L3). It is the depth conditioning (not intrinsics) that plays the larger role (L2 vs.\ L3), and removing it (L6) regresses 20/23 scenes it still completes, while \emph{fr1/rpy} diverges. Scale lowers the \textbf{Sim}(3) ATE further (0.032 to 0.026\,m) and cuts the unscaled \textbf{SE}(3) error from 0.364 to 0.034\,m and the map Chamfer distance from 6.6 to 3.1\,cm. The anchor is neutral on this panel (at most 4.9\,mm of \textbf{Sim}(3) ATE on any scene). Map-only keyframes keep trajectories identical while improving the Chamfer distance (3.27 to 3.05\,cm with $4.4\times$ the points). Hybrid mode gives comparable performance to the default mode, with a better median ATE, though scored per frame it is slightly worse (Sec.~\ref{sec:pose_estimation}).

\begin{table}[ht]
\centering
\caption{\textbf{Ablations}. Means and median ATE over the 24 scenes, Chamfer (cm) over the 15 map scenes, with \emph{stairs} separate. Rows L1--L5 add one component at a time up to the shipped default (def.); L6 and L7 leave out depth sites (23 scenes, as fr1/rpy diverges without cond). ref.: reference systems, not ranked (ORB-SLAM3 over its 21 completions). KF: keyframes.}
\label{tab:ablation}
{\scriptsize\setlength{\tabcolsep}{2pt}\renewcommand{\arraystretch}{0.95}\begin{tabular}{>{\raggedright\arraybackslash}p{2.45cm}rrrrr}
    \toprule
    \scriptsize \textbf{configuration} & \scriptsize \textbf{\shortstack[c]{Sim3\\mean\\(24)\,$\downarrow$}} & \scriptsize \textbf{\shortstack[c]{Sim3\\median\\(24)\,$\downarrow$}} & \scriptsize \textbf{\shortstack[c]{SE3\\mean\\(24)\,$\downarrow$}} & \scriptsize \textbf{\shortstack[c]{Chamfer\\(15)\,$\downarrow$}} & \scriptsize \textbf{stairs\,$\downarrow$} \\
    \midrule
    ORB-SLAM3 (ref., 21) & 0.0342 & 0.0178 & 0.0411 & -- & 0.045 \\
    MASt3R fresh (ref.) & 0.0304 & 0.0228 & 0.1104 & 4.35 & 0.016 \\
    L1 Pow3R, no priors & 0.0834 & 0.0449 & 0.4094 & 9.55 & 0.063 \\
    L2 + intrinsics & 0.0711 & 0.0431 & 0.4058 & 8.70 & 0.052 \\
    L3 + cond & 0.0319 & \underline{0.0189} & 0.3640 & 6.58 & 0.115 \\
    L4 + scale & \underline{0.0260} & 0.0191 & 0.0337 & \underline{3.07} & \underline{0.044} \\
    \textbf{L5 + anchor (def.)} & \textbf{0.0257} & 0.0191 & \underline{0.0329} & \textbf{3.05} & \underline{0.044} \\
    \midrule
    L6 def. $-$ cond (23) & 0.0577 & 0.0292 & 0.0658 & 5.56 & 0.066 \\
    L7 scale only (23) & 0.0547 & 0.0370 & 0.0705 & 5.40 & \textbf{0.033} \\
    \midrule
    map-only KF off & \textbf{0.0257} & 0.0191 & \underline{0.0329} & 3.27 & \underline{0.044} \\
    hybrid & 0.0264 & \textbf{0.0173} & \textbf{0.0314} & 3.15 & 0.062 \\
    \bottomrule
  \end{tabular}
}
\end{table}

We also study the sensitivity of our methods to depth quality. Scaling each valid depth pixel by $\max(1+\sigma\epsilon, 0.05)$, $\epsilon\sim\mathcal{N}(0,1)$, at all three sites raises the 24-scene mean ATE (0.026\,m unperturbed) monotonically, to 0.034, 0.056, 0.197 (one sequence diverging), and 0.609\,m for $\sigma = 0.05$, 0.1, 0.2, and 0.4. Noisy depth beats no depth (L2) at $\sigma = 0.1$, but not at 0.2. Keeping a random 50, 10, or 3\% of the depth pixels at the conditioning site instead gives 0.026, 0.028, and 0.026\,m, which are within 2.2\,mm of the unperturbed value. Thus, we show that the prior is robust to sparse depth, but not to dense, wrong depth. To test how depth image holes were impacting Pow3R-SLAM, we preprocessed the datasets with a depth completion network (Any2Full~\cite{zhou2026any-68d}). It regressed Replica ($+9.6\%$ mean ATE), was within noise on TUM ($-1.8\%$), and helped 7-Scenes ($-13.8\%$), but as it added compute cost it was not adopted. Completed pixels were less view-consistent than measured ones, and Pow3R already fills holes (Fig.~\ref{fig:Depthfig}).
\section{Limitations}\label{sec:limitations}
While the depth prior improves the prediction where sensor depth has holes, makes it metric, and improves aggregate results, it does not make the estimators more robust. Scenes for which Pow3R-SLAM is worse than MASt3R-SLAM include \emph{stairs} in 7-Scenes (Fig.~\ref{fig:stairs}), and \emph{planar\_3}, \emph{sofa\_2}, and \emph{repetitive} in ETH3D-SLAM. We hypothesize that repeating, self-similar structures cause these failures, as when Pow3R predicts a complete and confident geometry, the matching can find incorrect but mutually consistent correspondences that pull the backend to a wrong solution.

\section{Conclusions}
We presented Pow3R-SLAM, an RGB-D SLAM system that combines measured depth with learned two-view reconstruction to produce metric trajectories and dense maps. Depth conditions network predictions, sets pointmap scale, and anchors keyframe geometry during global optimization. Across 24 sequences from TUM, 7-Scenes, and Replica, Pow3R-SLAM runs $1.6\times$ faster than MASt3R-SLAM, with $15\%$ lower mean keyframe trajectory error and $3.1\times$ lower per-frame trajectory error without scale alignment. Its maps are denser, and achieve $30\%$ lower Chamfer distance on 7-Scenes and Replica. The hybrid ICP variant further improves runtime, reaching 25.3 FPS and a $2.1\times$ speedup over MASt3R-SLAM.

Ablations identify depth conditioning as the main source of accuracy gains and scaling as essential for metric output. Conditioning remains effective with only $3\%$ of depth pixels supplied to the network, showing that sparse measurements can still guide dense reconstruction. Depth noise and self-similar scenes remain challenging. Future work will investigate consistency-aware confidence gating, training with realistic sensor noise, and multi-view reconstruction priors. These results establish depth-conditioned pointmaps as a practical basis for real-time SLAM, combining improved trajectory accuracy and map completeness with lower runtime.

\addtolength{\textheight}{-9.5cm}

\section*{Acknowledgment}

We would like to thank the authors of MASt3R-SLAM for providing a strong and extensible baseline SLAM system. We would also like to thank the authors of Pow3R, for their strong work on integrating depth.

We acknowledge the support of the Natural Sciences and Engineering Research Council of Canada (NSERC).

AI Disclosure: Claude (Fable $5.1$, Opus $5$) was used to assist in development and debugging of the code for this work, including organizing the LaTeX for all Tables and Fig.~\ref{fig:pipeline}, and the Python code to stitch the images for Figs. \ref{fig:qual} and \ref{fig:stairs}. It was also used to help create the website and supplementary video. All output has been manually validated by the authors.

\bibliography{references_clean}

\begin{thebibliography}{10}
\providecommand{\url}[1]{#1}
\csname url@rmstyle\endcsname
\providecommand{\newblock}{\relax}
\providecommand{\bibinfo}[2]{#2}
\providecommand\BIBentrySTDinterwordspacing{\spaceskip=0pt\relax}
\providecommand\BIBentryALTinterwordstretchfactor{4}
\providecommand\BIBentryALTinterwordspacing{\spaceskip=\fontdimen2\font plus
\BIBentryALTinterwordstretchfactor\fontdimen3\font minus \fontdimen4\font\relax}
\providecommand\BIBforeignlanguage[2]{{%
\expandafter\ifx\csname l@#1\endcsname\relax
\typeout{** WARNING: IEEEtran.bst: No hyphenation pattern has been}%
\typeout{** loaded for the language `#1'. Using the pattern for}%
\typeout{** the default language instead.}%
\else
\language=\csname l@#1\endcsname
\fi
#2}}

\bibitem{newcombe2011kinectfusion:-793}
R.~A. Newcombe, A.~Fitzgibbon, S.~Izadi, O.~Hilliges, D.~Molyneaux, D.~Kim, A.~J. Davison, P.~Kohli, J.~Shotton, and S.~Hodges, ``{KinectFusion}: Real-time dense surface mapping and tracking,'' \emph{2011 10th {IEEE} International Symposium on Mixed and Augmented Reality}, pp. 127--136, 2011.

\bibitem{Whelan-RSS-15}
T.~Whelan, S.~Leutenegger, R.~F. Salas-Moreno, B.~Glocker, and A.~J. Davison, ``{ElasticFusion}: Dense {SLAM} without a pose graph,'' \emph{Robotics: Science and Systems}, 2015.

\bibitem{sturm2012benchmark-29f}
J.~Sturm, N.~Engelhard, F.~Endres, W.~Burgard, and D.~Cremers, ``A benchmark for the evaluation of {RGB-D} {SLAM} systems,'' \emph{2012 {IEEE}/{RSJ} International Conference on Intelligent Robots and Systems}, vol.~1, pp. 573--580, 2012.

\bibitem{wang2024dust3r:-980}
S.~Wang, V.~Leroy, Y.~Cabon, B.~Chidlovskii, and J.~Revaud, ``{DUSt3R}: Geometric {3D} vision made easy,'' \emph{2024 {IEEE}/{CVF} Conference on Computer Vision and Pattern Recognition ({CVPR})}, pp. 20\,697--20\,709, 2024.

\bibitem{leroy2024grounding-709}
V.~Leroy, Y.~Cabon, and J.~Revaud, ``Grounding image matching in {3D} with {MASt3R},'' \emph{European Conference on Computer Vision}, pp. 71--91, 2024.

\bibitem{jang2025pow3r:-27b}
W.~Jang, P.~Weinzaepfel, V.~Leroy, L.~Agapito, and J.~Revaud, ``{Pow3R}: Empowering unconstrained {3D} reconstruction with camera and scene priors,'' \emph{2025 {IEEE}/{CVF} Conference on Computer Vision and Pattern Recognition ({CVPR})}, pp. 1071--1081, 2025.

\bibitem{murai2025mast3r-slam:-690}
R.~Murai, E.~Dexheimer, and A.~J. Davison, ``{MASt3R}-{SLAM}: Real-time dense {SLAM} with {3D} reconstruction priors,'' \emph{2025 {IEEE}/{CVF} Conference on Computer Vision and Pattern Recognition ({CVPR})}, pp. 16\,695--16\,705, 2025.

\bibitem{mur-artal2015orb-slam:-3f2}
R.~Mur-Artal, J.~M.~M. Montiel, and J.~D. Tardós, ``{ORB-SLAM}: A versatile and accurate monocular {SLAM} system,'' \emph{{IEEE} Transactions on Robotics}, vol.~31, no.~5, pp. 1147--1163, 2015.

\bibitem{newcombe2011dtam:-406}
R.~A. Newcombe, S.~J. Lovegrove, and A.~J. Davison, ``{DTAM}: Dense tracking and mapping in real-time,'' \emph{2011 International Conference on Computer Vision}, vol.~1, pp. 2320--2327, 2011.

\bibitem{mur-artal2016orb-slam2:-44b}
R.~Mur-Artal and J.~D. Tardós, ``{ORB-SLAM2}: An open-source {SLAM} system for monocular, stereo, and {RGB-D} cameras,'' \emph{{IEEE} Transactions on Robotics}, vol.~33, no.~5, pp. 1255--1262, 2016.

\bibitem{campos2021orb-slam3:-eb8}
C.~Campos, R.~Elvira, J.~J.~G. Rodríguez, J.~M.~M. Montiel, and J.~D. Tardós, ``{ORB-SLAM3}: An accurate open-source library for visual, visual-inertial, and multimap {SLAM},'' \emph{{IEEE} Transactions on Robotics}, vol.~37, no.~6, pp. 1874--1890, 2021.

\bibitem{35402613541527}
Z.~Teed and J.~Deng, ``{DROID-SLAM}: deep visual {SLAM} for monocular, stereo, and {RGB-D} cameras,'' in \emph{Advances in Neural Information Processing Systems ({NeurIPS})}, 2021.

\bibitem{sucar2021imap:-1ff}
E.~Sucar, S.~Liu, J.~Ortiz, and A.~J. Davison, ``{iMAP}: Implicit mapping and positioning in real-time,'' \emph{2021 {IEEE}/{CVF} International Conference on Computer Vision ({ICCV})}, pp. 6209--6218, 2021.

\bibitem{zhu2022nice-slam:-578}
Z.~Zhu, S.~Peng, V.~Larsson, W.~Xu, H.~Bao, Z.~Cui, M.~R. Oswald, and M.~Pollefeys, ``{NICE-SLAM}: Neural implicit scalable encoding for {SLAM},'' \emph{2022 {IEEE}/{CVF} Conference on Computer Vision and Pattern Recognition ({CVPR})}, pp. 12\,776--12\,786, 2022.

\bibitem{yan2024gs-slam:-ec4}
C.~Yan, D.~Qu, D.~Xu, B.~Zhao, Z.~Wang, D.~Wang, and X.~Li, ``{GS-SLAM}: Dense visual {SLAM} with {3D} {Gaussian} splatting,'' \emph{2024 {IEEE}/{CVF} Conference on Computer Vision and Pattern Recognition ({CVPR})}, pp. 19\,595--19\,604, 2024.

\bibitem{wang2025vggt:-709}
J.~Wang, M.~Chen, N.~Karaev, A.~Vedaldi, C.~Rupprecht, and D.~Novotny, ``{VGGT}: Visual geometry grounded transformer,'' \emph{2025 {IEEE}/{CVF} Conference on Computer Vision and Pattern Recognition ({CVPR})}, pp. 5294--5306, 2025.

\bibitem{tolias2013aggregate-f56}
G.~Tolias, Y.~Avrithis, and H.~Jégou, ``To aggregate or not to aggregate: Selective match kernels for image search,'' \emph{2013 {IEEE} International Conference on Computer Vision}, pp. 1401--1408, 2013.

\bibitem{maggio2025vggt-slam:-22f}
D.~Maggio, H.~Lim, and L.~Carlone, ``{VGGT-SLAM}: Dense {RGB} {SLAM} optimized on the {SL}(4) manifold,'' \emph{Advances in Neural Information Processing Systems}, vol.~39, 2025.

\bibitem{liu2025slam3r:-64a}
Y.~Liu, S.~Dong, S.~Wang, Y.~Yin, Y.~Yang, Q.~Fan, and B.~Chen, ``{SLAM3R}: Real-time dense scene reconstruction from monocular {RGB} videos,'' \emph{2025 {IEEE}/{CVF} Conference on Computer Vision and Pattern Recognition ({CVPR})}, pp. 16\,651--16\,662, 2025.

\bibitem{tolias2020learning-198}
G.~Tolias, T.~Jenicek, and O.~Chum, ``Learning and aggregating deep local descriptors for instance-level recognition,'' \emph{European Conference on Computer Vision}, pp. 460--477, 2020.

\bibitem{chen1992object-868}
Y.~Chen and G.~Medioni, ``Object modelling by registration of multiple range images,'' \emph{Image and Vision Computing}, vol.~10, no.~3, pp. 145--155, 1992.

\bibitem{rusinkiewicz2001efficient-b64}
S.~Rusinkiewicz and M.~Levoy, ``Efficient variants of the {ICP} algorithm,'' in \emph{Third International Conference on 3D Digital Imaging and Modeling}, 2001.

\bibitem{glocker2013real-time-7bb}
B.~Glocker, S.~Izadi, J.~Shotton, and A.~Criminisi, ``Real-time {RGB-D} camera relocalization,'' \emph{2013 {IEEE} International Symposium on Mixed and Augmented Reality ({ISMAR})}, pp. 173--179, 2013.

\bibitem{straub2019replica-37a}
J.~Straub, T.~Whelan, L.~Ma, Y.~Chen, E.~Wijmans, S.~Green, J.~J. Engel, R.~Mur-Artal, C.~Ren, S.~Verma, A.~Clarkson, M.~Yan, B.~Budge, Y.~Yan, X.~Pan, J.~Yon, Y.~Zou, K.~Leon, N.~Carter, J.~Briales, T.~Gillingham, E.~Mueggler, L.~Pesqueira, M.~Savva, D.~Batra, H.~M. Strasdat, R.~D. Nardi, M.~Goesele, S.~Lovegrove, and R.~Newcombe, ``The {Replica} dataset: A digital replica of indoor spaces,'' \emph{{arXiv}}, 2019.

\bibitem{schöps2019bad-8c9}
T.~Schöps, T.~Sattler, and M.~Pollefeys, ``{BAD SLAM}: Bundle adjusted direct {RGB-D} {SLAM},'' \emph{2019 {IEEE}/{CVF} Conference on Computer Vision and Pattern Recognition ({CVPR})}, pp. 134--144, 2019.

\bibitem{schönberger2016structure-from-motion-314}
J.~L. Schönberger and J.-M. Frahm, ``{Structure-from-Motion} revisited,'' \emph{2016 {IEEE} Conference on Computer Vision and Pattern Recognition ({CVPR})}, pp. 4104--4113, 2016.

\bibitem{zhou2026any-68d}
Z.~Zhou, R.~Liu, T.~Liu, W.~Zuo, S.~Wang, Z.~Hong, and D.~Zhang, ``Any to full: Prompting {Depth Anything} for depth completion in one stage,'' \emph{{arXiv}}, 2026.

\end{thebibliography}

\end{document}